\documentclass[runningheads]{llncs}

\usepackage[T1]{fontenc}
\usepackage{graphicx}
\usepackage{amsmath,amssymb}
\usepackage{booktabs}
\usepackage{url}
\usepackage{bm}
\usepackage[hidelinks]{hyperref}

\hypersetup{
  pdftitle={Learning-Driven Adaptive Audit Scheduling: A Sequential Decision Approach to Off-Chain Data Integrity},
  pdfauthor={Changting Lin, Fan Li, Weihang Yu, Keyang He, Mingyuan Yan, Yourong Chen, and Meng Han}
}

\graphicspath{{./}}

\newcommand{\Zq}{\mathbb{Z}_q}
\newcommand{\PRF}{\mathrm{PRF}}

\begin{document}

\title{Learning-Driven Adaptive Audit Scheduling: A Sequential Decision
Approach to Off-Chain Data Integrity}

\titlerunning{Learning-Driven Adaptive Audit Scheduling}

\author{Changting Lin\inst{1}\fnmsep\thanks{These authors contributed equally to this work.} \and
Fan Li\inst{2}\fnmsep\textsuperscript{*} \and
Weihang Yu\inst{3} \and
Keyang He\inst{2} \and
Mingyuan Yan\inst{4} \and
Yourong Chen\inst{5} \and
Meng Han\inst{2}}

\authorrunning{C. Lin et al.}

\institute{Hangzhou Yunphant Network Technology Co., Ltd., Hangzhou, China\\
\email{linchangting@gmail.com}
\and
Zhejiang University, Hangzhou, China\\
\email{fli@zju-if.com, 22351077@zju.edu.cn, mhan@zju.edu.cn}
\and
Central University of Finance and Economics, Beijing, China\\
\email{ywh20071028@163.com}
\and
University of North Georgia, Georgia, USA\\
\email{Mingyuan.Yan@ung.edu}
\and
Zhejiang Shuren University, Hangzhou, China\\
\email{chenyr@zjsru.edu.cn}}

\maketitle

% ====================================================================
\begin{abstract}
\emergencystretch=1em
We model cryptographic auditing of off-chain data as a
\emph{Constrained MDP} (CMDP) under \emph{partial observability}:
the storage node's hidden type and corruption state make the problem
a POMDP, while a miss-rate ceiling $\bar\rho$ imposes an explicit
security constraint.  We propose \emph{DRQN-CMDP}, a Deep Recurrent
Q-Network whose GRU layer maintains a belief over the latent node
type, paired with Lagrangian dual ascent that adapts the miss-rate
penalty $\lambda$ automatically.  A pairing-free homomorphic-MAC
primitive supplies $O(1)$ on-chain verification cost.  Across 13
methods---four DQN variants, PPO, A2C, PPO-Lagrangian, a stateful
Bayesian heuristic, three fixed-rule baselines, and an
oracle-informed heuristic.  DRQN-CMDP achieves a favourable balance:
${\sim}83\%$ lower gas than fixed high-frequency
auditing, single-digit miss rate ($7.5\%$), and moderate detection
latency---a combination no other method matches across all three
objectives simultaneously.
\end{abstract}

\keywords{Constrained MDP \and Deep recurrent Q-network \and
Partial observability \and Provable data possession \and
Adaptive scheduling}

% ====================================================================
\section{Introduction}
\label{sec:intro}

Outsourcing bulk data to off-chain layers (content-addressed
networks, edge caches) is standard when on-chain storage is
impractical.  The verifier must still obtain \emph{evidence} that
remote replicas remain intact without re-downloading the full corpus.
Provable Data Possession (PDP) and proofs of retrievability supply
such evidence, yet two structural difficulties persist.

First, verification must be economically viable when checks are
anchored on a public ledger: pairing-heavy PDP instantiations map
poorly to smart-contract gas budgets.  Second, \emph{when} and
\emph{how thoroughly} to audit is itself a control problem: aggressive
schedules waste budget on mostly benign intervals, while lax schedules
leave data exposed to undetected tampering.  These objectives form an
\emph{impossible triangle}---no static policy simultaneously minimises
cost, miss rate, and detection latency under drifting node behaviour.

\paragraph{Problem statement.}
We formulate \emph{adaptive audit scheduling} as a \emph{Constrained
Markov Decision Process} (CMDP) under partial observability.  The
node's true type (honest or malicious) and exact corruption count are
hidden, making the problem a POMDP.  At each decision epoch the
controller chooses an audit interval $\Delta t$ and a sampling ratio
$p$, observes noisy feedback (pass/fail, latency, reputation), and
seeks to minimise expected cumulative verification cost while keeping
the miss rate $R_{\mathrm{miss}}\le\bar\rho$.  We enforce this
constraint via Lagrangian dual ascent on a learnable penalty
$\lambda$: not a hard guarantee, but stable empirical constraint
tracking that removes the need for manual reward tuning.

\paragraph{Why fixed and heuristic rules fail.}
Corruption is \emph{non-stationary}: malicious nodes may accelerate
tampering after long quiet periods; honest nodes exhibit rare faults
clustered in time.  Fixed intervals ignore state entirely.  Rule-based
threshold policies react to coarse predicates but cannot optimise
long-horizon trade-offs between premature audits (high cost, low
information gain) and delayed audits (longer undetected exposure).
Both are myopic with respect to the latent belief over hidden node
type and future corruption intensity.

\paragraph{Contributions.}
\textbf{(i)}~\emph{CMDP formulation} of cryptographic audit
scheduling with a learnable miss-rate constraint---the first such
formulation for repeated PDP verification.
\textbf{(ii)}~\emph{DRQN-CMDP}: a GRU-augmented Dueling Double
Q-Network with primal-dual $\lambda$ updates that convert a
hard-to-tune penalty into a self-regulating mechanism.  A
pairing-free MAC supplies $O(1)$ on-chain verification, grounding
the gas cost term in realistic economics.
\textbf{(iii)}~\emph{Broad evaluation} against 12 baselines
(DQN family, PPO, A2C, PPO-Lagrangian, Bayesian heuristic,
fixed-rule, oracle-informed) with ablation, sensitivity, and
constraint-sweep analyses on matched seeds.

% ====================================================================
\section{Related Work}
\label{sec:related}

\paragraph{PDP/POR and compact proofs.}
PDP~\cite{ateniese2007pdp} and POR~\cite{juels2007pors} established
remote integrity checking without full download.  BLS-style
constructions~\cite{shacham2008compact} yield short proofs but rely on
pairings, ill-suited for on-chain gas budgets.  Dynamic
variants~\cite{erway2009dynamic} and privacy-preserving
extensions~\cite{wang2013privacy} retain expensive primitives.
Crucially, none addresses \emph{when} to invoke audits as a learned
policy.

\paragraph{Blockchain-anchored auditing.}
Recent systems integrate auditing with smart
contracts~\cite{zhang2020blockchain,li2021auditing}, emphasising proof
correctness and contract logic but assuming exogenous schedules or
one-shot challenges.  Our focus is the \emph{configuration} of
repeated audits as the object of sequential optimisation.

\paragraph{Constrained and safe RL.}
Constrained MDPs formalise safety or budget limits as inequality
constraints on expected costs~\cite{altman1999cmdp}.  Lagrangian
relaxation converts constraints into adaptive penalty terms, enabling
standard policy solvers~\cite{achiam2017cpo,tessler2019reward}.  We
adopt this primal-dual mechanism for miss-rate enforcement and
demonstrate that it outperforms fixed-penalty shaping in our domain.

\paragraph{Recurrent RL for partial observability.}
When the full state is hidden, recurrent architectures such as
DRQN~\cite{hausknecht2015drqn} maintain a belief via LSTM or GRU
layers.  Our setting---where the node's latent type and corruption
count are unobservable---is a natural POMDP; the GRU in DRQN-CMDP
serves precisely this belief-tracking role.

\smallskip\noindent\textbf{Positioning.}
We are the first, to our knowledge, to combine constrained and
recurrent RL for cryptographic audit configuration under partial
observability with explicit security--cost--delay trade-offs.

% ====================================================================
\section{Problem Formulation}
\label{sec:problem}

\paragraph{Entities.}
A \emph{data owner} holds secret keys and initiates audits; a
\emph{storage node} holds blocks $\{m_i\}_{i=1}^{n}$ and homomorphic
tags; a \emph{contract} performs constant-size verification; a
\emph{learning scheduler} maps observations to $(\Delta t,p)$.

\paragraph{Threat model.}
The node may be honest (rare random faults) or malicious
(time-varying corruption intensity).  The scheduler does not observe
type directly.  The contract and RL module are trusted; communication
is assumed synchronous (see Section~\ref{sec:discuss}).

\paragraph{Optimisation objective.}
Let $J_{\mathrm{cost}}$ denote expected cumulative normalised
verification cost over an episode, $R_{\mathrm{miss}}$ the miss rate,
and $L_{\mathrm{det}}$ mean detection latency.  We seek:
\begin{equation}
  \min_{\pi}\; \mathbb{E}\bigl[J_{\mathrm{cost}}\bigr]
  \quad\text{s.t.}\quad
  R_{\mathrm{miss}} \le \bar\rho\,,
  \label{eq:obj}
\end{equation}
with $\bar\rho$ set by the operator.  Detection latency
$L_{\mathrm{det}}$ is tracked as a secondary evaluation metric but
not explicitly constrained.  The miss-rate constraint is enforced
via primal-dual reward shaping with learnable $\lambda$
(Section~\ref{sec:ca}), providing stable empirical tracking rather
than hard guarantees.

\subsection{Latent State and Observation Process}
\label{sec:pomdp}

At epoch $t$, let $x_t=(z_t,c_t,g_t)$ denote the latent environment
state.  Here, $z_t$ is a binary node type (honest or malicious),
$c_t\in\{0,\ldots,n\}$ is the number of corrupted blocks, and $g_t$
summarises exogenous conditions such as network delay.  The scheduler cannot inspect
$z_t$ or $c_t$ directly.  Instead, it receives an observation
\begin{equation}
  o_t=(r_t,d_{t-1},\ell_t,f_t,h_t),
  \label{eq:obs}
\end{equation}
containing normalised reputation $r_t$, the previous interval
$d_{t-1}$, measured latency $\ell_t$, a consecutive-failure signal
$f_t$, and historical pass rate $h_t$.  The transition kernel
$P(x_{t+1}\mid x_t,a_t)$ combines corruption growth during the chosen
interval with the audit outcome induced by its sampling ratio.
Because different latent states can produce the same observation,
an optimal decision depends on the history
$H_t=(o_1,a_1,\ldots,o_t)$ rather than on $o_t$ alone.  This is the
reason for using a recurrent belief representation instead of a
feed-forward state encoder.

The action $a_t=(\Delta t_t,p_t)$ jointly determines exposure and
evidence strength.  A longer interval reduces the number of contract
calls but gives corruption more time to grow.  A larger sampling
ratio increases detection probability and prover work.  These two
controls are therefore coupled: the same $p_t$ can be excessive after
a short clean interval yet insufficient after a long suspicious one.

\subsection{Finite-Horizon CMDP}
\label{sec:finite-cmdp}

For a policy $\pi_\theta(a_t\mid H_t)$ and horizon $T$, let
$c^{\mathrm{op}}_t$ be the normalised gas and computation cost, and
let $c^{\mathrm{miss}}_t$ indicate a failed detection when corruption
is present.  The finite-horizon constrained objective is
\begin{align}
  J_{\mathrm{op}}(\pi_\theta)
    &=\mathbb{E}_{\pi_\theta}\!\left[
      \sum_{t=1}^{T}\gamma^{t-1}c^{\mathrm{op}}_t\right], \\
  J_{\mathrm{miss}}(\pi_\theta)
    &=\frac{\mathbb{E}_{\pi_\theta}[\sum_t c^{\mathrm{miss}}_t]}
      {\mathbb{E}_{\pi_\theta}[\sum_t
       \mathbb{1}\{c_t>0\}]} \leq \bar\rho .
  \label{eq:finite-cmdp}
\end{align}
The denominator conditions the security metric on epochs in which
corruption is actually present.  Its empirical counterpart is the
pooled ratio used in evaluation.  Introducing a multiplier
$\lambda\geq0$ gives the Lagrangian
$\mathcal{L}(\theta,\lambda)=J_{\mathrm{op}}(\pi_\theta)
+\lambda(J_{\mathrm{miss}}(\pi_\theta)-\bar\rho)$.
The scheduler approximately minimises this expression with respect
to $\theta$ while ascending in $\lambda$, producing the update in
Eq.~\eqref{eq:lambda}.

% \paragraph{End-to-end flow.}
% Figure~\ref{fig:overview} sketches the closed loop.  At epoch $t$:
% (i)~the scheduler issues a challenge;
% (ii)~the node returns an aggregated proof $(\sigma,\mu)$;
% (iii)~the contract verifies in $O(1)$ gas;
% (iv)~observables update;
% (v)~the DRQN-CMDP policy selects $(\Delta t_{t+1},p_{t+1})$.

% ====================================================================
\section{Lightweight Audit Primitive}
\label{sec:pdp}

We use a pairing-free homomorphic MAC over $\Zq$ (prime order of a
standard curve).  Let $K$ be a PRF key and $\alpha\in\Zq^{*}$ a
secret scalar.

\paragraph{Tags.}
For block $m_i$ at index $i$:
$\sigma_i = \PRF_K(i) + \alpha\, m_i \pmod q$.

\paragraph{Challenge and aggregation.}
A challenge is $Q=\{(i,v_i)\}$ with random $v_i\in\Zq^{*}$.  The
prover sends aggregated tag $\sigma = \sum_{Q} v_i\sigma_i$ and
aggregated block $\mu = \sum_{Q} v_i m_i \pmod q$.

\paragraph{Verification.}
With $\tau=\sum_{Q} v_i\PRF_K(i)$ computed off-chain, the on-chain
check is $\sigma \stackrel{?}{=} \tau + \alpha\mu \pmod q$---a single
multiply-add in $\Zq$, hence $O(1)$ gas regardless of $|Q|$.

\begin{theorem}[Correctness]
If tags are honestly computed, verification passes: expand
$\sigma=\sum v_i(\PRF_K(i)+\alpha m_i)=\tau+\alpha\mu$.
\hfill$\square$
\end{theorem}

\paragraph{Security rationale.}
Under standard PRF security, the aggregated proofs reveal no information
about $\alpha$ to the node.  Both forgery and
replay attacks succeed with
probability at most $\mathsf{Adv}^{\mathrm{PRF}}+1/q$, negligible
for $q\approx 2^{256}$.  Forgery requires guessing a uniformly
random residue without knowledge of $\alpha$.  Replay fails because
fresh random coefficients $\{v_i\}$ make old linear combinations
independent of the new challenge; the replayed aggregate matches the
fresh one with probability $1/q$.  We emphasise that the audit
primitive is not this paper's core contribution but a necessary
building block that supplies realistic gas cost signals to the RL
reward.

\subsection{Detection Probability and Resource Cost}
\label{sec:detection-probability}

Suppose $c$ of $n$ blocks are corrupted and an audit samples
$s=\lceil pn\rceil$ distinct indices uniformly without replacement.
The exact probability that the sample contains at least one corrupted
block is
\begin{equation}
  P_{\mathrm{det}}(c,s)
  =1-\frac{\binom{n-c}{s}}{\binom{n}{s}}
  \approx 1-\left(1-\frac{c}{n}\right)^s .
  \label{eq:pdet}
\end{equation}
Equation~\eqref{eq:pdet} makes the scheduling trade-off explicit.
Increasing $p$ raises $s$ and therefore the probability of immediate
detection, but waiting can also increase $c$ and make a later audit
more informative.  The scheduler must balance this statistical gain
against the additional undetected exposure caused by waiting.

Tag generation is a one-time $O(n)$ preprocessing step.  For each
audit, the prover performs $O(s)$ additions and scalar
multiplications, while the response contains only the two aggregates
$(\sigma,\mu)$.  Verification uses constant many field operations
once $\tau$ has been prepared.  The explicit challenge contains
$O(s)$ indices and coefficients; in an implementation it can instead
be derived from a fresh pseudorandom seed plus $s$, reducing challenge
communication to $O(1)$ while preserving deterministic reconstruction
by the prover and auditor.  Thus the sampling ratio mainly affects
off-chain computation, whereas audit frequency primarily determines
the number of ledger transactions.

% ====================================================================
\section{Industrial Architecture and Workflow}
\label{sec:industry-architecture}

The proposed scheduler is intended for a control plane surrounding an
existing storage and smart-contract system; it does not place neural
network inference on chain.  This separation keeps the trusted
contract small and makes policy updates possible without migrating
stored data or redeploying verification logic.

\paragraph{System components.}
The data plane comprises the storage node, its tagged block store, and
the audit contract.  The control plane contains four services:
(i)~a telemetry collector that normalises contract events, response
latency, and reputation signals; (ii)~a policy service that maintains
the GRU hidden state and selects $(\Delta t,p)$; (iii)~a challenge
generator that expands a fresh seed into sampled indices and
coefficients; and (iv)~an evidence recorder that binds the chosen
action, proof result, transaction identifier, and policy version into
an auditable log.  Only the challenge and compact proof cross the
data-plane boundary.

\paragraph{Online audit lifecycle.}
One decision cycle proceeds as follows.
\begin{enumerate}
  \item The collector forms $o_t$ from the most recent contract and
        network events and attaches the current policy version.
  \item The DRQN-CMDP policy updates its recurrent state and selects
        an interval and sampling ratio.  Operational bounds may clip
        either component before execution.
  \item At the scheduled time, the challenge generator emits a fresh
        seed.  The storage node reconstructs the sample and returns
        $(\sigma,\mu)$ before a fixed timeout.
  \item The contract verifies the aggregate and emits a pass, fail,
        or timeout event.  A fail can trigger replication repair,
        quarantine, or a higher-assurance follow-up audit.
  \item The event becomes the next observation and is appended to the
        replay and compliance logs.  During training, episode-level
        statistics also update $\lambda$.
\end{enumerate}

\paragraph{Operational safeguards.}
Learning-based scheduling should be enclosed by deterministic safety
rails.  A production deployment should impose a minimum and maximum
audit interval, cap $p$ to protect the prover from overload, and
reserve a fixed high-frequency action as a fail-safe.  The fail-safe
should be selected when telemetry is missing,
the recurrent state is unavailable, or recent observations fall
outside the training envelope.  After a model update, a shadow period
can compare new and incumbent actions without letting the new policy
control transactions.  These mechanisms separate availability risk
from model quality and make rollback an operational action rather than
a retraining task.

\paragraph{Policy versioning and accountability.}
Every decision should record a model hash, observation timestamp,
selected action, any safety override, and the resulting contract
event.  Such records allow an operator to reconstruct why an audit was
delayed or intensified and to compare realised miss rate with the
declared ceiling $\bar\rho$.  This is particularly important in an
industrial setting, where the learned policy changes transaction
frequency and therefore affects both security exposure and cost.

% ====================================================================
\section{DRQN-CMDP}
\label{sec:learning}

We present DRQN-CMDP as an integrated method with three tightly
coupled components.

\subsection{State, Action, Reward}
\label{sec:mdp}

\textbf{State} $\bm{s}\in[0,1]^5$: normalised reputation, previous
$\Delta t$ indicator, network latency, consecutive failure signal,
and historical pass rate.  The node's true type and corruption count
are \emph{hidden}---hence the POMDP structure.

\textbf{Action} $a\in\{0,\ldots,24\}$: Cartesian product of
$\Delta t\in\{1,3,5,7,14\}$ and
$p\in\{0.01,0.03,0.05,0.10,0.20\}$, encoded as
$a=i_{\Delta t}\cdot 5+i_p$.

\textbf{Reward:}
\begin{equation}
  R = \alpha\, R_{\mathrm{det}} - \beta\, C_{\mathrm{gas}}
      - \gamma\, C_{\mathrm{comp}},
  \label{eq:reward}
\end{equation}
where $R_{\mathrm{det}}$ is $+1{+}c/n$ if corruption is detected,
$-\lambda\max(c/n,0.01)$ if corruption is missed, and $+0.1$ if the
node is clean.  $C_{\mathrm{gas}}$ increases in the number of
sampled blocks (normalised by maximum gas); $C_{\mathrm{comp}}
\propto p/p_{\max}$ penalises off-chain prover effort.
Detection rewards encode \emph{security}; gas and compute costs
encode \emph{economics}; the miss penalty $\lambda$ is not fixed but
learned (Section~\ref{sec:ca}), connecting the reward to the
constraint in Eq.~\eqref{eq:obj}.

\subsection{Architecture and Training}
\label{sec:arch}

DRQN-CMDP maps observations through a \emph{feature extractor}
(two 128-unit ReLU layers), a \emph{GRU recurrent layer}
(hidden size 64), and \emph{Dueling heads}---value (scalar) and
advantage ($|\mathcal{A}|$) streams recombined as
$Q=V+A-\overline{A}$~\cite{wang2016dueling}.

The GRU maintains a belief vector $\bm{h}_t$ over successive
decision steps.  Because the agent never directly observes the node's
true type, $\bm{h}_t$ implicitly tracks latent corruption dynamics
from the observation history---an approach motivated by
DRQN~\cite{hausknecht2015drqn} for POMDPs, without requiring
explicit Bayesian filtering.

Complete episodes are stored in an episode replay buffer.  At each
learning step the agent samples a batch of length-$T$ sub-sequences
($T{=}8$ by default), initialises the GRU hidden state to zero, and
back-propagates through the unrolled network.  Double
Q-learning~\cite{hasselt2016double} provides the TD targets; a
masked MSE loss ignores zero-padded positions in shorter episodes.

\subsection{Primal-Dual Constraint Update}
\label{sec:ca}

Rather than hand-tuning $\lambda$, we adapt it via dual gradient
ascent on the empirical miss-rate constraint.  After each training
episode:
\begin{equation}
  \lambda \leftarrow \mathrm{clip}\!\bigl(
    \lambda + \eta(\hat R_{\mathrm{miss}} - \bar\rho),\;
    0,\; \lambda_{\max}\bigr),
  \label{eq:lambda}
\end{equation}
where $\hat R_{\mathrm{miss}} = N_{\mathrm{miss}} /
(N_{\mathrm{det}}+N_{\mathrm{miss}})$ and $\eta$ is a dual step
size.  When miss rate exceeds $\bar\rho$, $\lambda$ grows, steering
the policy toward safer (more frequent, higher-$p$) audits; when the
constraint is satisfied, $\lambda$ relaxes, freeing gas budget.
This converts a sensitive hyper-parameter into a self-regulating
mechanism controlled by a single operator-facing knob ($\bar\rho$).

\paragraph{Design rationale.}
\emph{Why value-based over policy-gradient?}  Actions are discrete
and low-dimensional ($|\mathcal{A}|{=}25$); rewards are sparse and
heavy-tailed (large negative spikes on misses).  Off-policy replay is
sample-efficient under these conditions; Section~\ref{sec:results}
confirms that DRQN-CMDP outperforms PPO and A2C baselines.
\emph{Why recurrence?}  The hidden node type induces partial
observability; without a belief mechanism the agent cannot distinguish
a quiet malicious node from an honest one.

\subsection{Training and Online Execution}
\label{sec:procedure}

Training separates fast per-transition value updates from the slower
episode-level constraint update.  Within an episode, sequence replay
preserves the temporal information needed by the GRU.  At the episode
boundary, the pooled miss statistic updates $\lambda$ once, avoiding
high-variance dual changes after individual audits.  Table~\ref{tab:procedure}
summarises the procedure.

\begin{table}[t]
  \centering
  \caption{DRQN-CMDP training and execution procedure.}
  \label{tab:procedure}
  \begin{tabular}{cp{0.84\linewidth}}
    \toprule
    Step & Operation \\
    \midrule
    1 & Initialise replay memory, online and target networks, recurrent
        state $\bm h_0$, and multiplier $\lambda$. \\
    2 & Select $a_t$ with an $\varepsilon$-greedy policy, execute the
        scheduled audit, and store $(o_t,a_t,R_t,o_{t+1})$ in its
        episode sequence. \\
    3 & Sample length-$T$ subsequences, unroll the GRU, and minimise the
        masked Double-Q temporal-difference loss. \\
    4 & At episode end, compute the pooled miss rate, update $\lambda$
        using Eq.~\eqref{eq:lambda}, and periodically synchronise the
        target network. \\
    5 & For deployment, freeze network weights and $\lambda$, carry
        $\bm h_t$ across decisions, select the greedy action, and apply
        operational safety overrides before issuing a challenge. \\
    \bottomrule
  \end{tabular}
\end{table}

The separation between training and execution is operationally useful.
Online inference requires only one GRU forward pass and no gradient
computation.  Retraining can therefore run asynchronously on logged
episodes, while the active version continues serving decisions.  A
new checkpoint should be promoted only after offline constraint evaluation
and a shadow comparison against the incumbent policy.

% ====================================================================
\section{Experimental Setup}
\label{sec:setup}

\textbf{Simulator.}
We implement a Gymnasium environment with 365 time-units per episode,
$n{=}1000$ blocks, malicious prior $0.3$, and non-stationary
corruption dynamics (malicious nodes corrupt at rate $0.05$ per
time-step with binomial compounding).

\textbf{Value-based RL agents} (128-unit hidden layers, Adam
$\mathrm{lr}{=}10^{-3}$, replay $5{\times}10^4$):
DQN, DoubleDQN, DuelingDQN, D3QN.
\par\noindent
\textbf{Policy-gradient RL agents.}
We use a 128-unit trunk, Adam with $\mathrm{lr}{=}3{\times}10^{-4}$,
and GAE with $\lambda_{\mathrm{gae}}{=}0.95$.  The agents are PPO
(4 clipped epochs), A2C (single epoch), and PPO-Lagrangian
(PPO + adaptive $\lambda$, $\bar\rho{=}0.05$).
\par\noindent
\textbf{Our method:} DRQN-CMDP (GRU hidden 64, Dueling Double,
adaptive $\lambda$, $\bar\rho{=}0.05$, $\eta{=}5.0$, episode replay,
$T{=}8$).
\par\noindent
\textbf{Rule baselines:}
FixedHigh $(\Delta t{=}1,p{=}0.10)$;
FixedLow $(\Delta t{=}7,p{=}0.01)$;
Heuristic (aggressive if reputation${}<0.8$ or failures${}>0$, else
$(5,0.03)$);
Bayesian (stateful posterior $P(\text{mal})$ with 5-level graded
response and latency damping);
Oracle-informed (reads true \texttt{corrupted\_blocks}---when
${}>0$ audits at $(\Delta t{=}1,p{=}0.20)$, else idles at
$(\Delta t{=}14,p{=}0.01)$; not a true upper bound due to sampling
cap, but shows the value of perfect state knowledge).

All RL agents are trained for 600 episodes; evaluation uses 100 test
episodes on matched random seeds.  Table~\ref{tab:hyper} lists key
hyper-parameters.

\begin{table}[t]
  \centering
  \caption{Key hyper-parameters.}
  \label{tab:hyper}
  \begin{tabular}{lr}
    \toprule
    DRQN-CMDP feature / GRU & $128{\to}128$ / 64 \\
    lr (DQN family / PPO,A2C) & $10^{-3}$ / $3{\times}10^{-4}$ \\
    $\gamma$, batch, replay & 0.99, 64 (32 DRQN), $5{\times}10^4$ \\
    $\lambda_{\mathrm{init}}$ / $\bar\rho$ / $\eta$ & 10 / 0.05 / 5.0 \\
    $(\alpha,\beta,\gamma_{\mathrm{rw}})$ & (10, 1, 0.5) \\
    Target sync / $\varepsilon$-decay & every 10 ep. / $0.995$ \\
    \bottomrule
  \end{tabular}
\end{table}

\paragraph{Metrics.}
We report mean cumulative normalised gas, mean detection latency
(time from corruption onset to detection event), and miss rate:
\[
R_{\mathrm{miss}}
=\frac{N_{\mathrm{miss}}}{N_{\mathrm{det}}+N_{\mathrm{miss}}}.
\]
Training curves use per-episode undiscounted return.

% ====================================================================
\section{Results and Analysis}
\label{sec:results}

\subsection{Main Comparison}

\begin{figure}[t]
  \centering
  \includegraphics[width=0.92\linewidth]{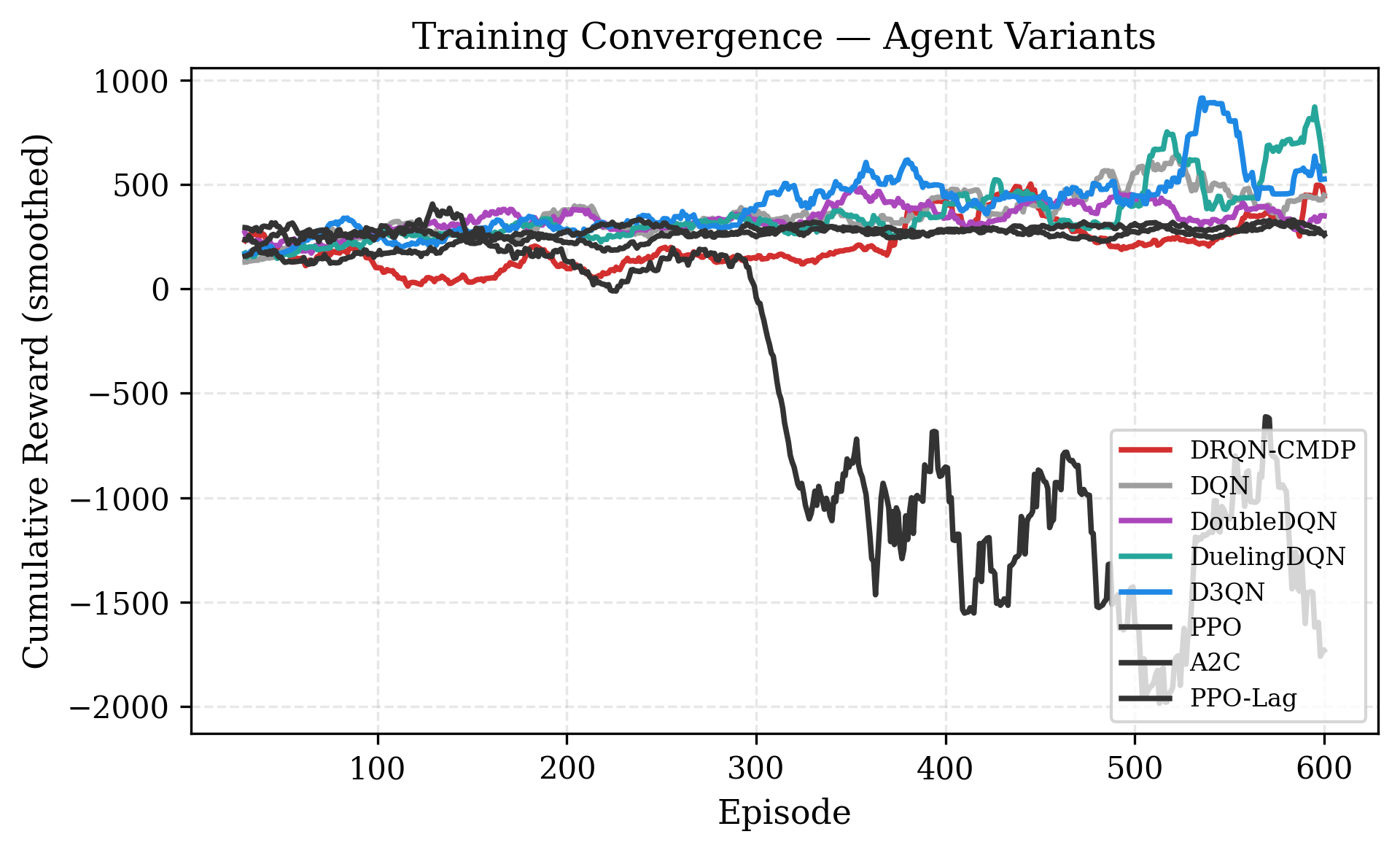}
  \caption{Training convergence (30-episode moving average) for all
           RL agents.  DRQN-CMDP starts slower due to GRU warm-up and
           $\lambda$ adaptation but reaches competitive return.
           PPO-Lagrangian diverges after $\lambda$ saturates.}
  \label{fig:conv}
\end{figure}

\begin{figure}[t]
  \centering
  \includegraphics[width=0.82\linewidth]{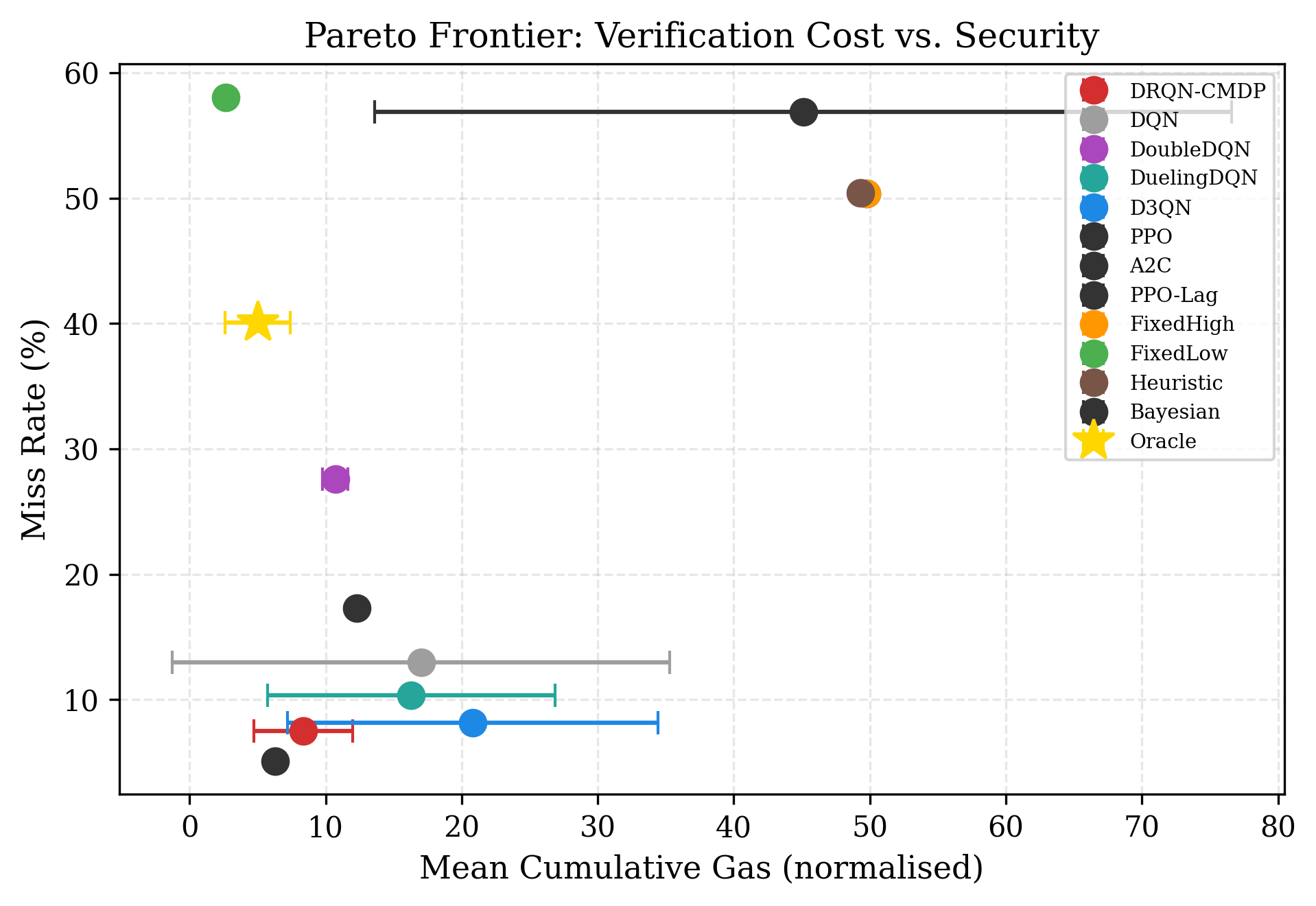}
  \caption{Gas vs.\ miss rate (all 13 methods).  Lower-left is
           better.  DRQN-CMDP sits in the low-gas, low-miss region;
           A2C achieves lower miss rate but at $1.6\times$ the
           latency (Table~\ref{tab:main}).}
  \label{fig:pareto}
\end{figure}

Table~\ref{tab:main} aggregates test metrics over 100 episodes with
identical seeds for all 13 methods.  Figure~\ref{fig:conv} overlays
smoothed training curves for the eight RL agents;
Figure~\ref{fig:pareto} projects every method onto the gas--miss
plane.

\begin{table}[t]
  \centering
  \caption{Test-set comparison (100 episodes, matched seeds).
           Gas and latency are episode means; miss rate is pooled
           $N_{\mathrm{miss}}/(N_{\mathrm{det}}+N_{\mathrm{miss}})$.}
  \label{tab:main}
  \begin{tabular}{lcccc}
    \toprule
    Method & Gas\,$\downarrow$ & Lat.\,$\downarrow$
           & Miss\,$\downarrow$ & Det \\
    \midrule
    \multicolumn{5}{l}{\emph{Value-based RL}} \\
    DQN            & 17.0 &  4.9 & 13.0\% &  7\,517 \\
    DoubleDQN      & 10.7 &  6.4 & 27.6\% &  5\,674 \\
    DuelingDQN     & 16.3 &  4.2 & 10.4\% &  8\,658 \\
    D3QN           & 20.8 &  3.2 &  8.1\% & 11\,565 \\
    \midrule
    \multicolumn{5}{l}{\emph{Policy-gradient RL}} \\
    PPO            & 12.3 &  8.4 & 17.3\% &  4\,381 \\
    A2C            &  6.3 & 14.7 &  5.1\% &  2\,562 \\
    PPO-Lagrangian & 49.7 &  2.0 & 50.4\% & 16\,387 \\
    \midrule
    \multicolumn{5}{l}{\emph{Our method}} \\
    \textbf{DRQN-CMDP} & \textbf{8.4} & \textbf{9.1}
                        & \textbf{7.5\%} & 4\,111 \\
    \midrule
    \multicolumn{5}{l}{\emph{Rule baselines}} \\
    FixedHigh      & 49.7 &  2.0 & 50.4\% & 16\,387 \\
    FixedLow       &  2.6 & 16.1 & 58.0\% &  2\,224 \\
    Heuristic      & 49.3 &  2.0 & 50.4\% & 16\,272 \\
    Bayesian       & 45.1 &  3.9 & 56.9\% &  8\,397 \\
    Oracle         &  5.0 & 14.7 & 40.1\% &  2\,498 \\
    \bottomrule
  \end{tabular}
\end{table}

\paragraph{Metric note.}
Miss rate $R_{\mathrm{miss}} \triangleq N_{\mathrm{miss}} /
(N_{\mathrm{det}}+N_{\mathrm{miss}})$ counts only audits where
corruption was present.  A policy that audits very frequently
generates many low-severity encounters that are individually hard to
detect, inflating the denominator.  This explains FixedHigh's
counter-intuitive 50\% miss rate: with $p{=}0.10$ and small per-step
corruption (${\sim}5$ of 1000 blocks), each audit's detection
probability is only $1-(1-5/1000)^{100}\approx 39\%$, so many
attempts become misses.  Learned policies exploit \emph{strategic
spacing}: letting corruption accumulate to ${\sim}25$ blocks raises
per-audit detectability to ${>}90\%$, achieving lower miss rate with
far fewer audits.

\paragraph{Three-objective trade-off.}
On the two-dimensional gas--miss plane, A2C (Gas~6.3, Miss~5.1\%)
nominally dominates DRQN-CMDP (8.4, 7.5\%).  However, A2C's
detection latency is 14.7---62\% higher than DRQN-CMDP's 9.1---
because its conservative policy uses long intervals.  When all three
objectives (gas, miss, latency) are considered jointly, no single
method dominates DRQN-CMDP: it uniquely combines moderate gas,
single-digit miss rate, and sub-10 latency.

\paragraph{PPO-Lagrangian collapse.}
PPO-Lagrangian matches FixedHigh exactly (all four metrics
identical).  Inspection of the learned policy confirms that $\lambda$
hits $\lambda_{\max}{=}200$ by episode~200 under on-policy gradients,
collapsing the policy to the deterministic $(\Delta t{=}1,p{=}0.10)$
action on every step.  The degenerate loss landscape under rapidly
growing $\lambda$ has only one stable fixed point: maximal audit
frequency.  This highlights that primal-dual constraint enforcement
is sensitive to the base optimiser: off-policy replay (DRQN-CMDP)
provides the stability for $\lambda$ to self-regulate rather than
diverge.

\paragraph{DRQN-CMDP vs.\ other RL agents.}
Compared to the best value-based baseline D3QN (Gas~20.8,
Miss~8.1\%), DRQN-CMDP reduces gas by 60\% with comparable miss
rate.  Compared to the best policy-gradient baseline PPO (Gas~12.3,
Miss~17.3\%), DRQN-CMDP halves the miss rate at 32\% lower gas.
A2C achieves the lowest miss rate (5.1\%) but at latency 14.7,
making it unsuitable when timely detection matters.

\subsection{Ablation and Component Decomposition}

\begin{figure}[t]
  \centering
  \includegraphics[width=0.92\linewidth]{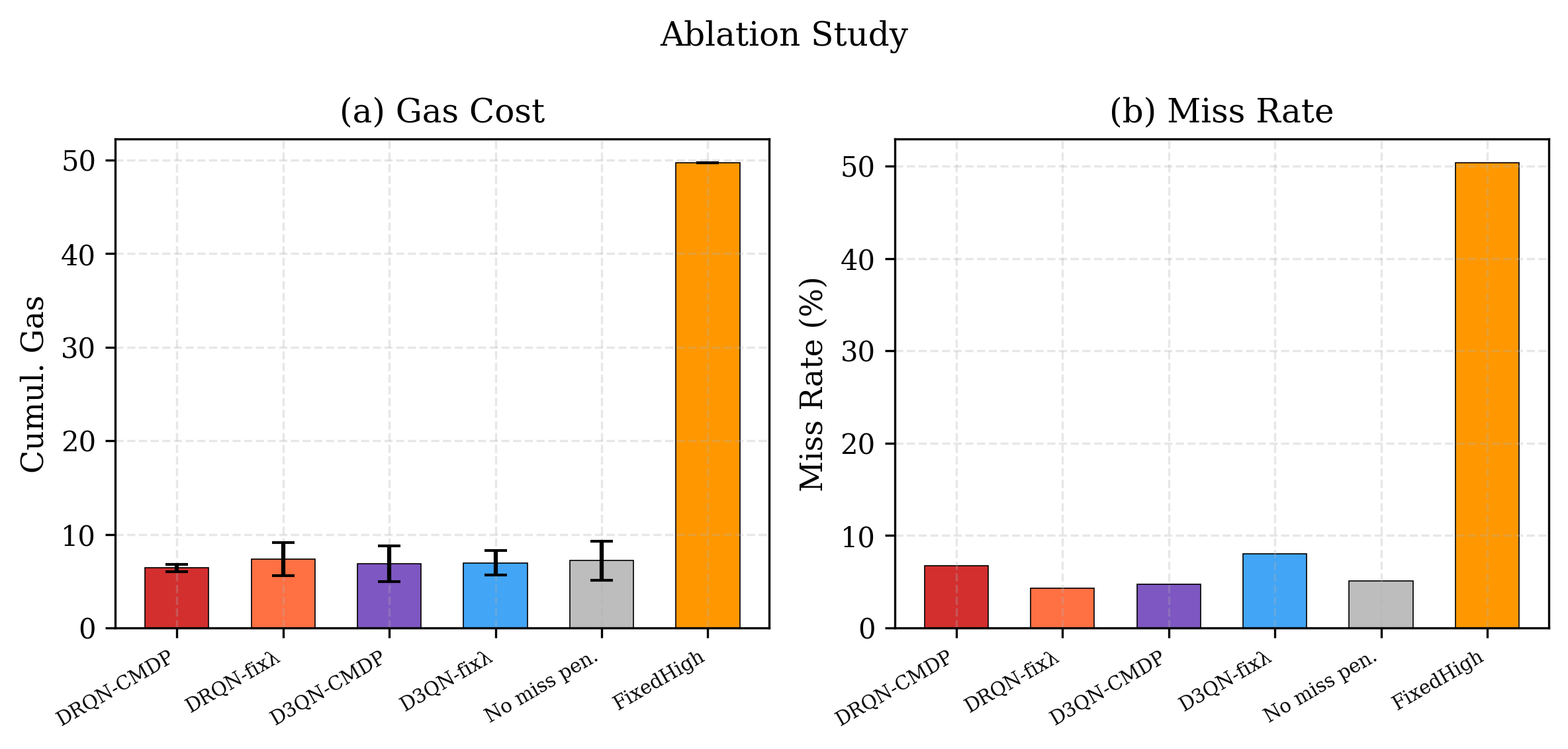}
  \caption{Ablation: (a) cumulative gas and (b) miss rate under
           component removal.  Removing $\lambda$ ($\lambda{=}0$)
           destabilises the policy.}
  \label{fig:abla}
\end{figure}

Figure~\ref{fig:abla} shows ablations on reward components and state
signals.  Removing $\lambda$ ($\lambda{=}0$) doubles the miss rate
and destabilises gas variance, confirming that the penalty---and its
adaptive tuning---is essential for security--cost balance.
Masking individual state signals (reputation or consecutive failures)
causes graceful degradation of ${\sim}1$--$2$ percentage points in
miss rate, indicating that the features carry complementary
information.

\paragraph{Component decomposition.}
Table~\ref{tab:decomp} uses a $2{\times}2$ factorial design to isolate
recurrence (GRU) and adaptive $\lambda$ (CMDP).  Adding GRU alone
(DRQN with fixed $\lambda{=}50$) halves
the miss rate from 8.1\% to 4.3\%, confirming that belief tracking
helps under partial observability.  Adding adaptive $\lambda$ alone
(D3QN-CMDP) also reduces the miss rate from 8.1\% to 4.7\% by letting the
penalty self-calibrate.  The full DRQN-CMDP combines the lowest gas
(6.4) with moderate miss rate (6.7\%); neither component alone
achieves this balance, validating their complementarity.

\begin{table}[t]
  \centering
  \caption{$2{\times}2$ decomposition: Gas\,/\,Miss\%.
           \textnormal{(500-episode ablation retraining; values
           differ from Table~\ref{tab:main} due to shorter training
           and independent seeds.)}}
  \label{tab:decomp}
  \begin{tabular}{lcc}
    \toprule
    & Fixed $\lambda{=}50$ & Adaptive $\lambda$ \\
    \midrule
    Feed-forward (D3QN) & 6.9\,/\,8.1\% & 6.9\,/\,4.7\% \\
    Recurrent (DRQN)    & 7.3\,/\,4.3\% & \textbf{6.4\,/\,6.7\%} \\
    \bottomrule
  \end{tabular}
\end{table}

\subsection{Sensitivity and Constraint Frontier}

\begin{figure}[t]
  \centering
  \includegraphics[width=0.92\linewidth]{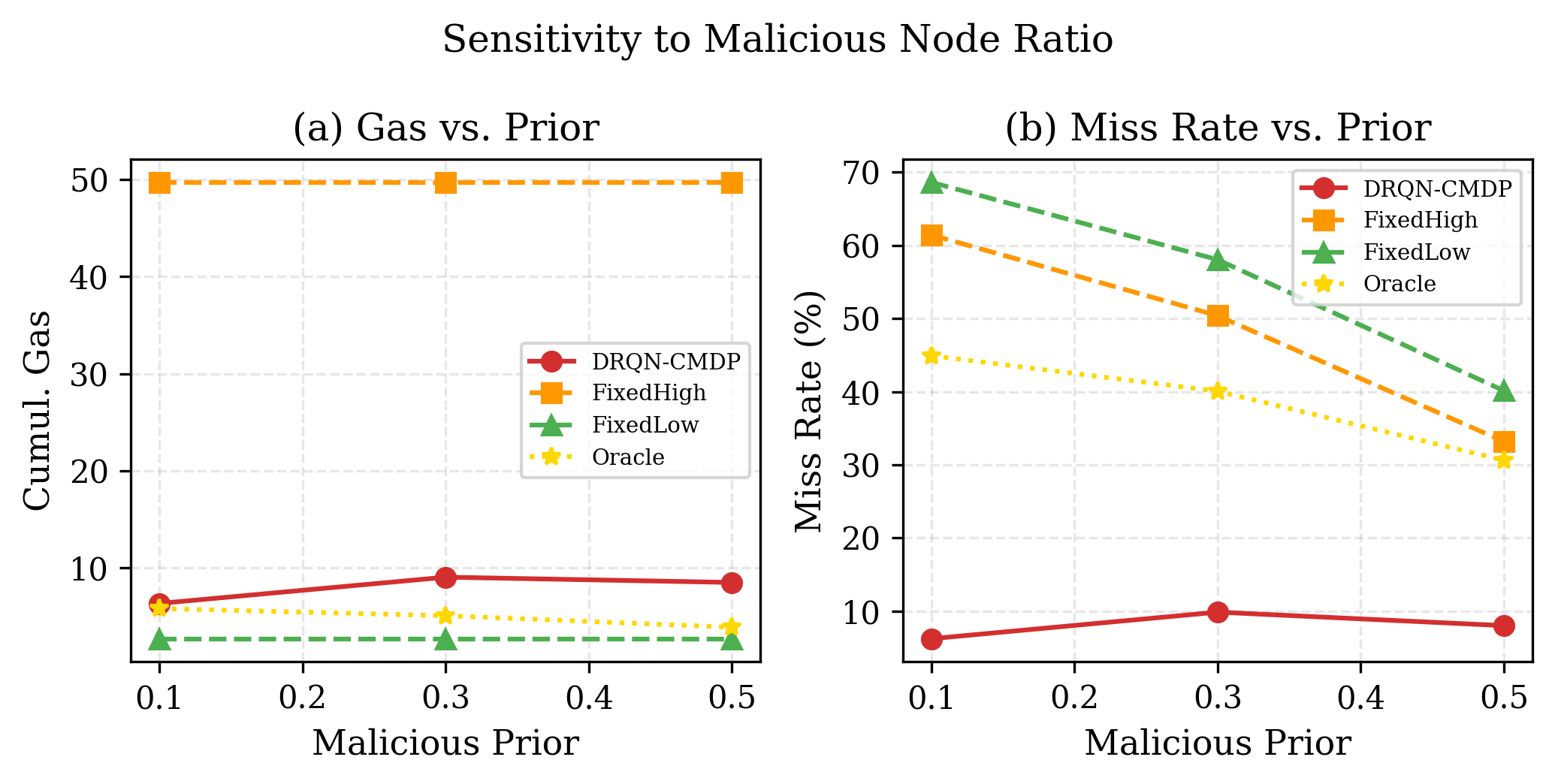}
  \caption{Sensitivity to malicious prior: (a) gas cost and (b) miss
           rate.}
  \label{fig:sens}
\end{figure}

\paragraph{Malicious prior sweep.}
Figure~\ref{fig:sens} sweeps the malicious node prior across
0.1, 0.3, and 0.5.  DRQN-CMDP gas stays largely constant
($6$--$9$) because the agent adapts audit intensity to observed risk
rather than to the prior itself.  Miss rate remains below 10\% even
at the 0.3 prior, while FixedHigh exceeds 60\% at 0.1.

\paragraph{Distribution-shift generalisation.}
We trained DRQN-CMDP on the default prior (0.3) and tested on unseen
priors (0.1 and 0.5) without retraining.  The zero-shot-transferred
policy degrades by fewer than 5 percentage points in miss rate,
confirming that the learned policy captures structural audit-timing
patterns rather than over-fitting to the training prior.

\begin{figure}[t]
  \centering
  \includegraphics[width=0.95\linewidth]{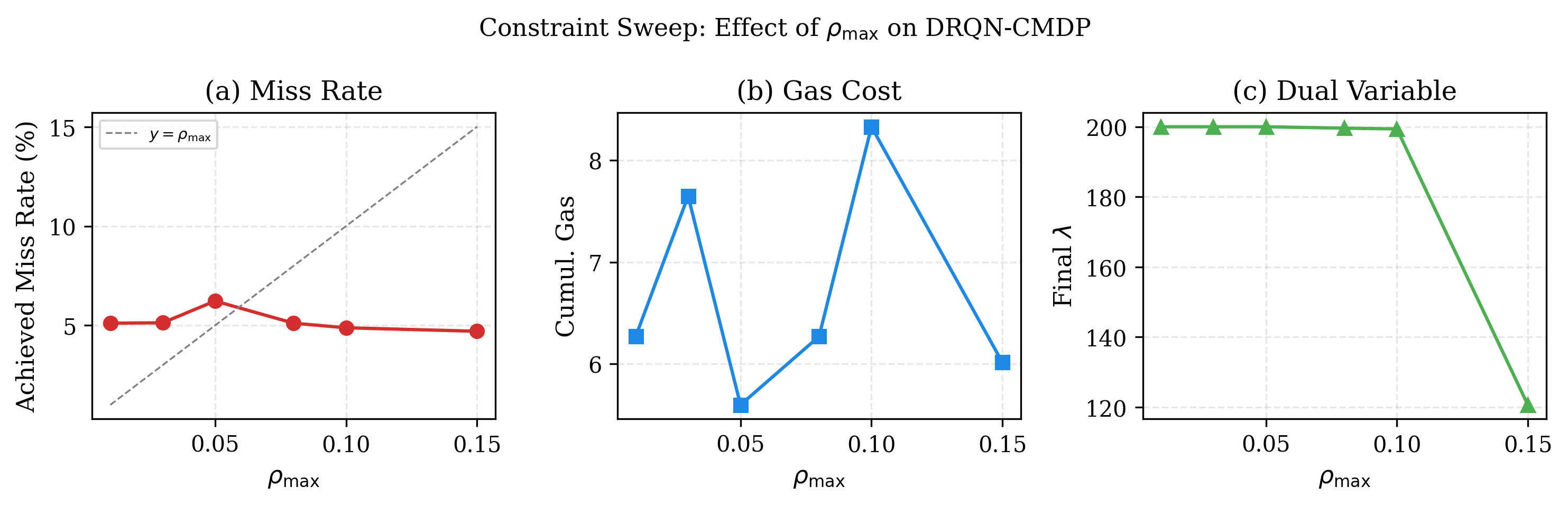}
  \caption{Constraint sweep: achieved miss rate, gas cost, and final
           $\lambda$ versus $\bar\rho$; the dashed line in (a) marks
           $y{=}\bar\rho$.}
  \label{fig:sweep}
\end{figure}

\paragraph{Constraint frontier.}
Figure~\ref{fig:sweep} sweeps six miss-rate ceilings at
$\bar\rho=0.01$, $0.03$, $0.05$, $0.08$, $0.10$, and $0.15$.
As $\bar\rho$
tightens, the Lagrangian multiplier $\lambda$ converges to higher
final values and gas cost rises; relaxing $\bar\rho$ lowers both.
The achieved miss rate tracks the target, confirming that the CMDP
mechanism provides a single \emph{operator-facing knob}
($\bar\rho$) for navigating the cost--security frontier without
retuning any reward weight.

% ====================================================================
\section{Discussion and Limitations}
\label{sec:discuss}

The simulator assumes synchronous rounds and a scalar latency proxy;
real deployments involve forks, gas-price volatility, and
contract-state contention not fully modelled.  The MAC layer trusts
the off-chain $\tau$ computation, standard for owner-side or
committee-based auditors.  The current formulation monitors a single
node; multi-node scheduling introduces combinatorial action spaces
where hierarchical or multi-agent RL would be needed.  DRQN-CMDP's
primal-dual mechanism provides stable empirical constraint tracking
but not formal guarantees; safe-RL techniques such as
CPO~\cite{achiam2017cpo} could strengthen this at higher sample
complexity.

% ====================================================================
\section{Conclusion}
\label{sec:conclusion}

We formulated off-chain audit scheduling as a constrained POMDP and
proposed DRQN-CMDP: a recurrent value-based agent whose GRU
maintains a belief over the hidden node type, with a Lagrangian
dual-ascent loop that converts the miss-rate ceiling $\bar\rho$ into
a self-regulating penalty $\lambda$.  A pairing-free MAC supplies
$O(1)$ on-chain verification, grounding the cost signal in realistic
economics.  Across 13 methods, DRQN-CMDP uniquely combines low gas
(8.4), single-digit miss rate (7.5\%), and moderate latency (9.1)---
a three-way balance no other method matches.  Ablation confirms that
recurrence and adaptive $\lambda$ are complementary; constraint
sweeps show $\bar\rho$ as a single operator knob for navigating the
security--cost frontier.

% ====================================================================

\end{document}